\documentclass[letterpaper, 10 pt, journal, twoside]{ieeetran}
\IEEEoverridecommandlockouts                              % This command is only needed if
\usepackage[pdftex]{graphicx}
\usepackage{amsmath} % assumes amsmath package installed
\usepackage{amssymb}  % assumes amsmath package installed
\usepackage{subfigure}
\usepackage{multirow}
\usepackage{array,booktabs}
\usepackage{diagbox}
\usepackage{balance}
\usepackage{nccmath}
\makeatletter
\let\NAT@parse\undefined
\makeatother
\usepackage[numbers]{natbib}

\usepackage{irap_SIunits}
\usepackage{irap_acronyms}
\usepackage{irap_math}
\usepackage{irap_misc}

\usepackage{soul,color}

\usepackage{xcolor}

\usepackage{lipsum}

\usepackage{caption}

\title{\LARGE \bf
Line as a Visual Sentence: \\
Context-aware Line Descriptor for Visual Localization
}

\author{Sungho Yoon${}^{1}$ and Ayoung Kim${}^{2*}$% <-this % stops a space
\thanks{Manuscript received: April 29, 2021; Revised August 2, 2021; Accepted September 7, 2021. This paper was recommended for publication by Editor Sven Behnke upon evaluation of the Associate Editor and Reviewers' comments. This work was fully supported by [Localization in changing city] project funded by Naver Labs Corporation.} %Use only for final RAL version
\thanks{$^{1}$S. Yoon is with the Robotics Program,
        KAIST, Daejeon, S. Korea {\tt\footnotesize sungho.yoon@kaist.ac.kr}}%
\thanks{$^{2}$A. Kim is with the Department of Mechanical Engineering, SNU, Seoul, S. Korea {\tt\footnotesize ayoungk@snu.ac.kr}}
\thanks{Digital Object Identifier (DOI): see top of this page.}
}

\begin{document}
%\onecolumn
\maketitle

\begin{abstract}
Along with feature points for image matching, line features provide additional constraints to solve visual geometric problems in robotics and \ac{CV}. Although recent \ac{CNN}-based line descriptors are promising for viewpoint changes or dynamic environments, we claim that the \ac{CNN} architecture has innate disadvantages to abstract variable line length into the fixed-dimensional descriptor. In this paper, we effectively introduce Line-Transformers dealing with variable lines. Inspired by \ac{NLP} tasks where sentences can be understood and abstracted well in neural nets, we view a line segment as a sentence that contains points (words). By attending to well-describable points on a line dynamically, our descriptor performs excellently on variable line length. We also propose line signature networks sharing the line's geometric attributes to neighborhoods. Performing as group descriptors, the networks enhance line descriptors by understanding lines' relative geometries. Finally, we present the proposed line descriptor and matching in a \ac{PL-Loc}. We show that the visual localization with feature points can be improved using our line features. We validate the proposed method for homography estimation and visual localization.
\end{abstract}

% Keywords appear just beneath the abstract. Use only for final RAL version.

\section{Introduction}
\label{sec:intro}

\IEEEPARstart{V}{isual} features are widely utilized in many robotics and \acf{CV} applications such as visual tracking, \ac{SLAM}, and \ac{SFM}. While the keypoint features are well-studied and a base of many applications, keypoints that are not well distributed in the image may result in unstable and inaccurate pose estimation.

Recent research has reported that \ac{SLAM} performance can be enhanced by using both points and lines \cite{Gomez2016, Pumarola2017, Gomez2019, Yang2019} even for low-textured environments. For example, \ac{LBD}~\cite{LBD2013} is one of the widely-used line descriptors in \ac{SLAM}. Reliable performance of \ac{LBD} has been reported for continuous frames, whereas the performance degrades under a wide baseline, which prevented the line-based approach from adapting line features directly in visual localization~\cite{Toft2020, Paul2019}. Tackling this limitation, approaches leveraged \acf{CNN} to learn the representation of line descriptors \cite{LLD2019, DLD2019, WLD2020} yielding superior performance. Still, \ac{CNN} has innate difficulties in dealing with variable line lengths because the dimension of \ac{CNN} should be predefined by its architecture. Another issue of line occlusion is also related to this length variation and was examined in SOLD$^2$~\cite{Pautrat2021}.

% FIGURE
\begin{figure}[!t]
	\centering
	\includegraphics[width=0.99\columnwidth]{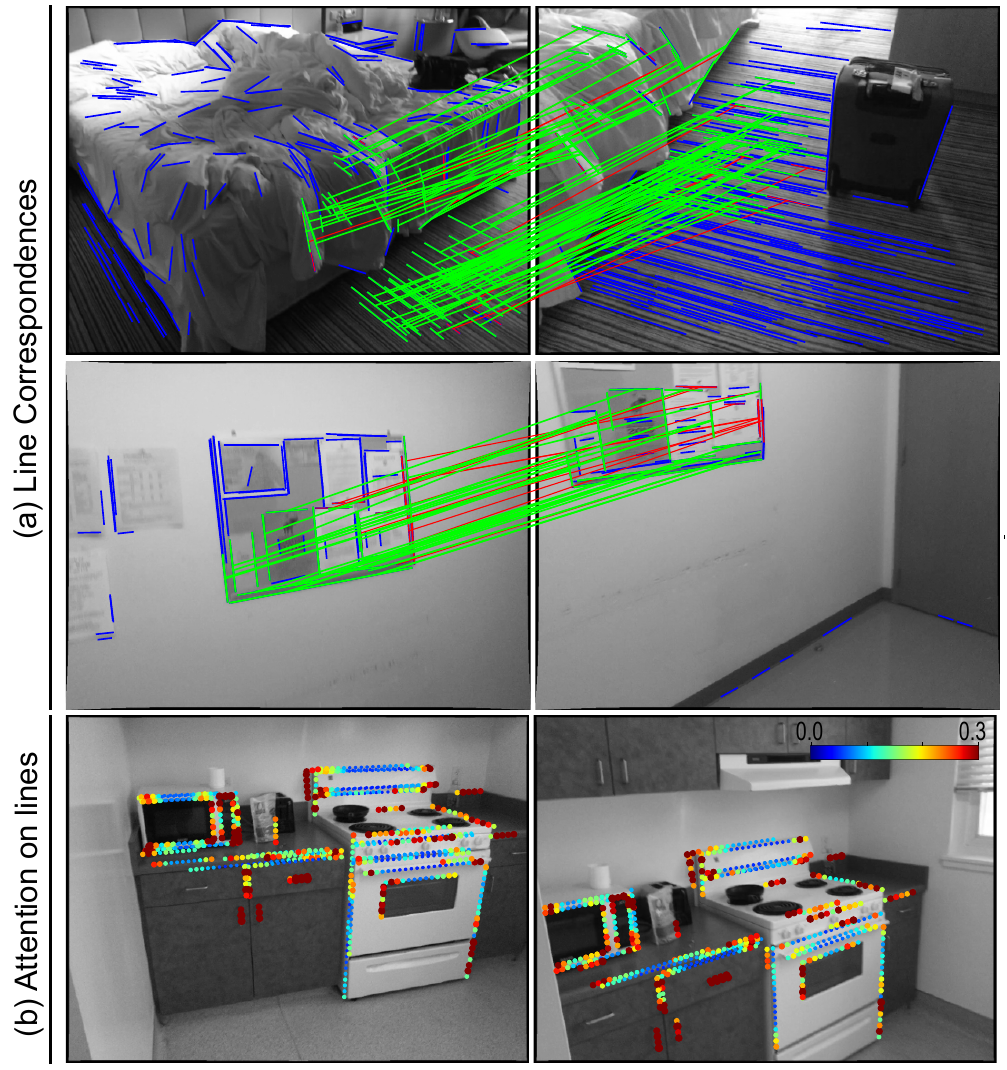}
	\caption{(a) For both low and high-textured environments, our line descriptor performs accurate matching. (b) Attention scores for points in a line with matched lines having similar attentional contexts.}
	\label{fig:main}
	\vspace{-5mm}
\end{figure}
% FIGURE

In our work, we overcome this fixed length requirement and achieve flexibility to the length variation by interpreting a line segment from the \acf{NLP} point of view. We view a line consisting of points as a sentence consisting of words, thereby allowing various lengths for lines. The word2vec in \ac{NLP} learns vector representations that encapsulate each distinct word, leveraging them as base inputs of RNN, LSTM, or transformers for later tasks such as text classification and text summarization. Similarly, we utilize a learned descriptor map as a transition from RGB pixels to point vectors.

Based on this key idea, we propose a novel line descriptor using transformers~\cite{Vaswani2017}. The transformers in our model learn the context of the point vectors within a line segment to summarize it in the form of a line descriptor. Unlike SOLD$^2$, a line is not just a simple sequence of points but is handled with attention to its context dynamically, enabling reliable performance even under occlusion.
%We validate that the proposed method achieves state-of-the-art performance in homography estimation and visual localization. We present a \acf{PL-Loc} for a visual localization using the proposed line descriptor and matching.
The attributes of the proposed method can be summarized below.
%Our experiment shows the best localization performance can be achieved using point and line features together in a camera pose optimization.

% FIGURE
%\begin{figure}[!t]
%	\def\width{1\columnwidth}%
%	\centering
%	\includegraphics[width=1.0\width]{figures/main.pdf}
%	\caption{Line-Transformers: a learnable line descriptor for visual localization. We introduce a novel line descriptor to represent a line context using a transformer architecture.}
%	\label{fig:main}
%	\vspace{-5mm}
%\end{figure}
% FIGURE

%----------------------------------------------------------------------------%

% FIGURE: flowchart
\begin{figure*}[!t]
	\centering
	\includegraphics[width=0.8\textwidth]{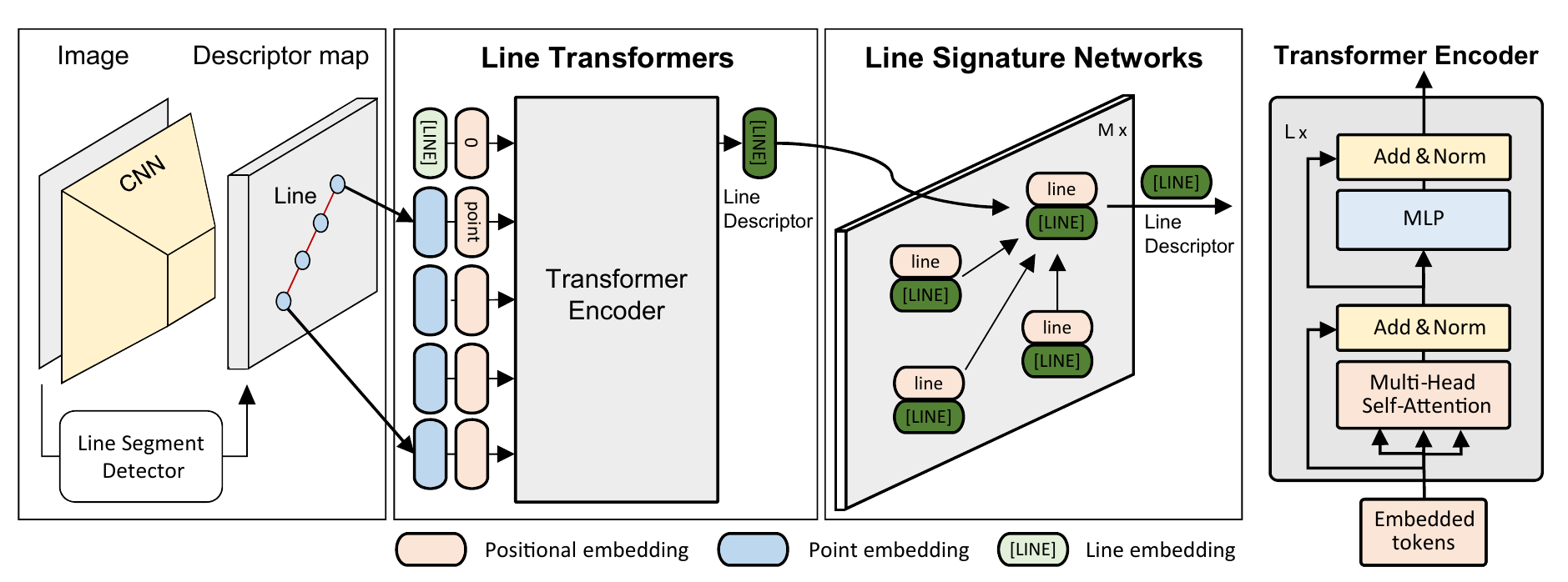}
	\caption{Line-Transformers consist of two major components: \textit{line transformers} and \textit{line signature networks}. The first component uses a \textit{line tokenizer} to extract point tokens and embeddings from a line segment. Considering the context of the point embeddings, transformers summarize it into a line embedding, or a line descriptor. The second component enhances the line descriptor by sharing lines' positional context with its neighborhoods.}
	\label{fig:overview}
	\vspace{-5mm}
\end{figure*}
% figure

\begin{itemize}
	% As we've seen in \ac{NLP} tasks, we utilize point descriptors on a line segment to summerize as a line context.
	\item We present a novel line segment descriptor using a transformer architecture by considering line segments as sentences and points as words. Leveraging \ac{NLP} for line descriptor, we can handle lines with various lengths.

	\item The proposed line descriptor understands the line segment's context by paying attention to more meaningful points on a line. It effectively abstracts various length lines to a fixed-sized descriptor.

	\item We suggest line signature networks that share the message of line attributes (e.g., position, angle, length, and descriptors) between neighborhoods. Serving as a grouped line descriptor, the line descriptors can learn geometric attributes of their neighborhoods.

%	\item We complete a \ac{PL-Loc} pipeline that achieves accurate and robust performance. We show that the proposed line descriptor can boost localization performance together with keypoints using a Perspective-n-Point and Line (PnPL) optimization. As a result, the PL-Loc outperforms the SuperPoint-based localization by 36.3\% in the indoor environments. % 0.25m{\slash}10\textdegree
\end{itemize}
\section{Related Work}
\label{sec:related_work}

%----------------------------------------------------------------------------%
\subsection{Line Descriptors}

%Line matching approaches can be classified into two types: individual line descriptors and the methods using the geometric relationships between points or line segments nearby.

As a handcrafted line descriptor, \citeauthor{MSLD2009}~\cite{MSLD2009} proposed a \ac{MSLD} observing gradients around line segments. \ac{LBD}~\cite{LBD2013} enhanced the precision and computing time by investigating gradients on bands that are parallel to a line. Recently, \citeauthor{LLD2019}~\cite{LLD2019} presented a \ac{LLD} for visual \ac{SLAM} in a deep-learning manner. They first built a full-sized descriptor map using \ac{CNN} and uniformly split a line segment into a fixed-number of subsegments. \citeauthor{DLD2019} proposed a learning-based line descriptor named DLD~\cite{DLD2019}. They also split a line segment and trained the \ac{CNN} with a self-generated dataset. Extended works from the same group~\cite{WLD2020} adapted wavelets to extract more meaningful information from an image, referred to as the \ac{WLD}. \citeauthor{Pautrat2021}~\cite{Pautrat2021} introduced SOLD$^2$ for joint line detection and description. To find a line correspondence, SOLD$^2$ sampled uniformly-spaced point descriptors on a line and performed the sequence matching.

%----------------------------------------------------------------------------%
\subsection{Point Descriptors}

The recent point descriptor also focused on learning-based methods. Like a handcrafted feature SIFT~\cite{Lowe2004}, learned descriptors often utilized the patch as their inputs as in L2-Net~\cite{Tian2017} and HardNet~\cite{Mishchuk2017}. Examining keypoint detection and description together, LIFT~\cite{Yi2016} introduced an end-to-end learning pipeline implementing keypoint detection, orientation estimation, and description simultaneously. SuperPoint~\cite{DeTone2017} proposed self-supervised learning for detector and descriptor using synthetic pretraining and homography adaptation.
%D2-Net~\cite{Dusmanu2019} presented a describe-and-detect approach, which detects reliable keypoints tightly-coupled with the feature descriptor so more suitable in aspect to feature matching. R2D2\cite{Revaud2019} considers the descriptor's discriminativeness together with jointly learning keypoint detection and description so it achieved more robust feature matching. More recently, DISK\cite{Tyszkiewicz2020} proposed a policy gradient method of reinforcement learning for keypoint selection and descriptor matching.

%----------------------------------------------------------------------------%
%\subsection{Line and point matching}

%\hl{cos distance point matching superglue?}

%\hl{Too short? organization? need? no matching for points} Besides, line matching methods are also studied utilizing geometric relationships between points and line segments\cite{Jia2018, Fan2012} or between lines and line segments\cite{Li2016, Wang2009}. % Since the methods are considering more than two lines or points together to match line segments, we do not consider in this paper.

%----------------------------------------------------------------------------%
\subsection{Point-and-Line \ac{SLAM}}

Exploiting both point and line for \ac{SLAM} has strong advantages, robustly performing in challenging environments such as low-textured, motion blurred, and defocused cases~\cite{Gomez2016, Pumarola2017, Gomez2019, Yang2019}. PL-SVO~\cite{Gomez2016} extended the SVO method by adapting its semi-direct approach to line segments. PL-SLAM~\cite{Pumarola2017} presented line features on the monocular ORB-SLAM. In \cite{Gomez2019}, stereo camera-based PL-SLAM was presented to improve the bag-of-words method using points and lines.

In this paper, we utilize visual localization to evaluate line descriptors. Visual localization~\cite{Toft2020, Paul2019} is a problem of estimating a camera pose given a query image and a prior 3D feature map, which is similar to relocalization in \ac{SLAM}. It differs from the previous works in considering only discrete scenes, including large viewpoint changes, and it is effective for evaluating the robustness of line descriptors in various situations.
%Besides, several point and line-based \ac{SLAM}s are also studied with intertial sensors\cite{Yang2019, Yijia2018}.

%----------------------------------------------------------------------------%
\subsection{Transformers in \ac{NLP} and \ac{CV}}

We briefly summarize recent studies connecting NLP and CV. \citeauthor{Vaswani2017} proposed the transformer for language translation~\cite{Vaswani2017}, and it became a base architecture of many state-of-the-art methods in \ac{NLP} tasks. \ac{BERT} is one of the widely used models utilizing transformer encoders~\cite{Devlin2018}. One of the recent trends in \ac{CV} is the adoption of the transformers~\cite{Dosovitskiy2020, Wu2020}. \citeauthor{Dosovitskiy2020} proposed a Vision Transformer (ViT)~\cite{Dosovitskiy2020} in which 16x16 patches from an image are used as inputs of a standard transformer for image classification. Visual Transformer (VT)~\cite{Wu2020} adopted a transformer using semantic tokens from convolutions and spatial attention so as to treat an image focussing more on important regions.
%It pre-trained the model with unlabeled text to learn the bi-directional context of sentences, and it achieved state-of-the-art performance for various \ac{NLP} tasks by small fine-tuning with an additional layer.

%----------------------------------------------------------------------------%
\section{Method}
\label{sec:method}

The proposed Line-Transformers aim to build a line descriptor given points located on a line segment (\figref{fig:overview}).

% illustrates the overall architecture.
%In contrast to the previous approaches, we use a transformer architecture that understands the language context successfully in recent research

% FIGURE: sentence and line
\begin{figure}[!b]
	\centering
	\includegraphics[width=0.75\columnwidth]{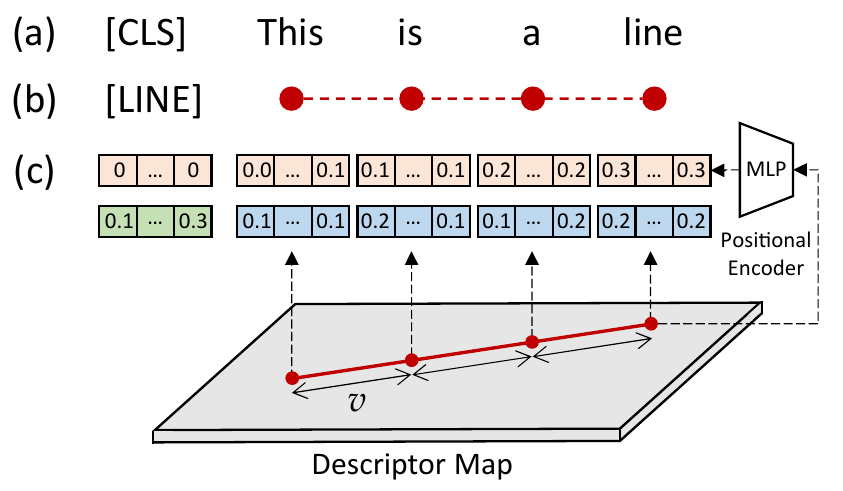}
	\caption{(a) In BERT~\cite{Devlin2018}, a sentence is tokenized, and a \texttt{[CLS]} is prepended for a sentence classification task. (b) Similar to BERT, a line segment is tokenized, and a special line token \texttt{[LINE]} is used for aggregating points information. (c) The point embedding is a descriptor vector located on each point token on a descriptor map.}
	\label{fig:sentence}
	\vspace{-5mm}
\end{figure}
% figure

%============================================================================%
\subsection{Line Transformers}

%----------------------------------------------------------------------------%
\subsubsection{Line Tokenizer}

We briefly illustrate \ac{NLP} terminologies and explain our method based on a similar concept. In \ac{NLP}, word tokenization is a process used to divide a sentence into smaller words called tokens. The tokens are utilized as minimum units for model inputs. Special tokens are also used to perform special tasks. For example, the classification token \texttt{[CLS]} is applied to aggregate sequence representation for classification tasks and the separator token \texttt{[SEP]} is used to differentiate two sentences. These tokens are converted into vector representations by allocating words with similar meaning into a similar vector space. The vector representation of tokens is called word embedding, and \ac{NLP} models exploit them to understand natural languages effectively. In our problem, we formulate a point-and-line segment as the relationship between a word and a sentence in the natural language.

As in \figref{fig:sentence}, the line tokenizer aims to generate point tokens and their embeddings that describe a line segment. After detecting line segments from an image~\cite{Gioi2010}, we uniformly select points $\bvec{p}$ on a line segment. The position of point $\bvec{p}$ consists of its image coordinates and keypoint confidence: $\bvec{p_i} = (x,y,c)_i$. The inter-points interval ($v$) is then determined depending on the level of discriminability. When the interval of points is small, the model can receive more information to describe a line but needs larger computations and memory. The number of points ($n$) on a line is $n = \lfloor\ell/v\rfloor+1$ when a line length is $\ell$. As the \texttt{[CLS]} token in BERT~\cite{Devlin2018}, we introduced a special line token \texttt{[LINE]} to aggregate contextual information of point tokens. The line token is prepended on the series of point tokens.

We encode each point token into a vector representation, point embedding~$\bvec{E}\in\mathbb{R}^{1 \times D}$, where $D$ is the descriptor dimension. This is done by associating each point token with a vector that matched in a dense descriptor map~\cite{DeTone2017} encoded with a descriptor vector at each pixel. The embedding of the special token \texttt{[LINE]}, $\bvec{E_{line}}\in\mathbb{R}^{1 \times D}$, expresses the initial state of a line descriptor, and its weight values are learned during the training process. Finally, we add the embeddings to the positional embeddings~$\bvec{E_{pos}\in\mathbb{R}^{(n+1) \times D}}$ which is obtained by \ac{MLP} using each point's position.

%----------------------------------------------------------------------------%
\subsubsection{Transformer}

Given the token embeddings, we model a line descriptor using transformers~\cite{Vaswani2017}. The transformer encoder is composed of two sublayers: \ac{MSA} layers and \ac{MLP} layers, whereas each sub-layer has a residual connection and \ac{LN}.

We stack the transformer $L$ times as in \eqref{eq:transformer}. Here, the token embeddings~$\bvec{z_0}$ serve as the encoder inputs. The line embedding~$\bvec{E_{line}}$ is located at the first element of $\bvec{z_0}$, and is expressed with the superscript $0$. Then, at the last layer~$\bvec{z_{\textit{L}}}$, the line embedding contains the line context (i.e., $\bvec{E_{line}=\bvec{z_0^0}}$ and a line descriptor $\bvec{d} = \bvec{z_{\textit{L}}^0}$).

\begin{eqnarray}
  \label{eq:transformer}
  \bvec{z_0} &=& [\bvec{E_{line}}; \bvec{E_1}; \bvec{E_2}; ... ;\bvec{E_{\textit{n}}}] + \bvec{E_{pos}},  \\
  \bvec{z'_{\textit{i}-1}} &=& \text{LN}(\text{MSA}(\bvec{z_{\textit{i}-1}}, \bvec{mask_0}) + \bvec{z_{\textit{i}-1}}), \nonumber \\
  \bvec{z_{\textit{i}}} &=& \text{LN}(\text{MLP}(\bvec{z'_{\textit{i}-1}}) + \bvec{z'_{\textit{i}-1}}), \textit{i} = 1...\textit{L} \nonumber \\
  \bvec{d} &=& \bvec{z_{\textit{L}}^0} \nonumber
\end{eqnarray}

Each line segment has a various number of point tokens based on the length $\ell$. To handle the various sizes at once, we use a mask vector $\bvec{mask_0}$ for the \ac{MSA} to reject the unrelated point embeddings inside of scaled dot-product attention. The mask has the size $1 \times n_{max}$.

%============================================================================%
\subsection{Line Signature Networks}

Inspired by line signatures~\cite{Wang2009} and message-passing~\cite{Sarlin2019, Velickovic2018} in \ac{GNN}, the proposed deep line signature networks are designed for line signatures' messages to be shared with each line segment using graph attention networks. The line signature is originally proposed as a grouped line descriptor. It clusters neighbor line segments as a group and takes relative positions by a series of angles and length ratios. While the line signature needs to define the clustering range of neighbor line segments and the man-defined attributes of lines, we exploit a graph attention network that can implicitly assign neighbor line segments and pass the messages of their descriptions including positions.

\begin{eqnarray}
  \label{eq:line_signature}
  \bvec{d'_{\textit{i}}} &=& \bvec{d_{\textit{i}}} + \text{MLP}(\bvec{x_{\textit{i}}},\bvec{y_{\textit{i}}},\ell_{\textit{i}},\cos\theta_{\textit{i}},\sin\theta_{\textit{i}}) \\
  \bvec{s_0} &=& [\bvec{d'_1};\bvec{d'_2};...;\bvec{d'_{\textit{m}}}] \nonumber \\
  \bvec{s_{\textit{j}}} &=& \bvec{s_{\textit{j}-1}} + \text{MLP}\Big(\bvec{s_{\textit{j}-1}} || \text{MSA}(\bvec{s_{\textit{j}-1}})\Big), \textit{j} = 1...\textit{M} \nonumber
\end{eqnarray}

We first make an attribute embedding by feeding the line's attributes such as midpoints $(x,y)$, angle $\theta$, and length $\ell$ to \ac{MLP}. Then, we add it on descriptor $\bvec{d_i}$ and share the messages with each line segment using the message-passing network as in \eqref{eq:line_signature}. The operator $||$ represents concatenation, and $m$ denotes the number of line descriptors within an image. We also stack the message-passing layers $M$ times. Finally, we normalize line descriptors in \bvec{z_{\textit{M}}} after feeding them to another \ac{MLP}.

% FIGURE
\begin{figure}[!t]
	\centering
	\includegraphics[width=0.7\columnwidth]{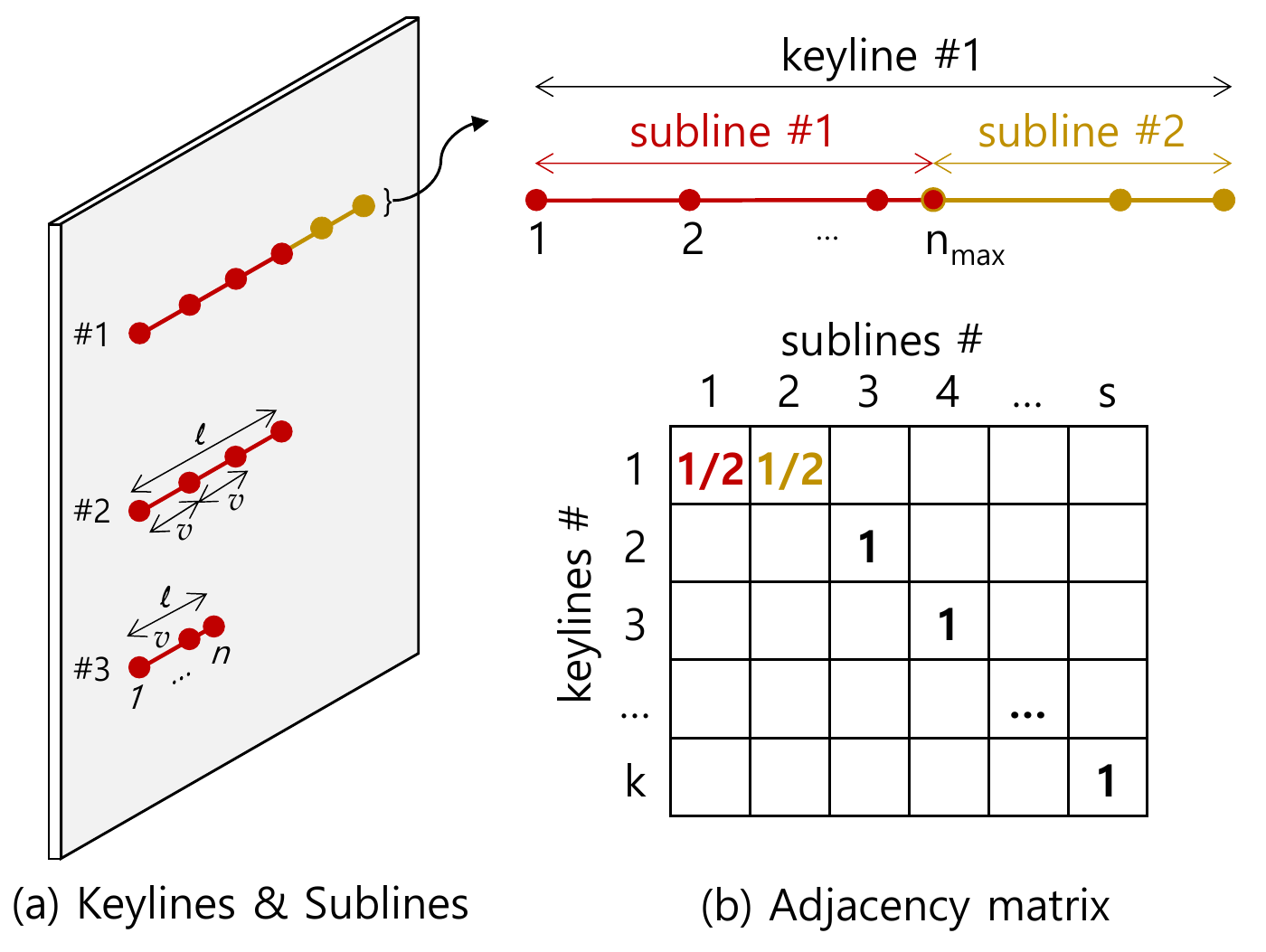}
	\caption{Keylines, sublines, and adjacency matrix. Points are selected with the interval $v$ on a line segment. When the number of points $n$ is greater than $n_{max}$, the keyline is divided into two sublines as in (b). The adjacency matrix describes relationship between keylines and sublines.}
  \label{fig:adjacency_matrix}
	\vspace{-5mm}
\end{figure}
% FIGURE

%============================================================================%
\subsection{Sublines to Keylines}

For transformers, the number of input tokens is limited by its maximum size ($n_{max}$). Overcoming the limited text length of transformers, the longer sequences were often truncated in NLP.
%\citeauthor{Sun2020}~\cite{Sun2020} investigated to handle the long texts using various truncation methods and hierarchical methods, and they experimentally chose to use a head+tail truncation method selecting the first 128 tokens and the last 382 tokens.
Instead of truncation, we handle long line segments utilizing the keyline and subline concepts. Let us call the original line segment a \textit{keyline}. When the keyline is longer than the maximum~($\ell > n_{max} \times v$), we divide it into multiple \textit{sublines}. In doing so, we can build an adjacency matrix~$\mat{A_{img}}$ via the relationship between keylines and sublines. The values of the adjacency matrix are one divided by the number of sublines, as presented in \figref{fig:adjacency_matrix}(b). Then, the distance matrix of sublines $\mat{C_{sublines}}$ can be transformed into a distance matrix of keylines $\mat{C_{keylines}}$ as follows:
\begin{equation*}
  \label{eq:subline2keyline}
  \mathbf{\mat{C_{keylines}} = \mat{A_{img1}} \cdot \mat{C_{sublines}} \cdot \mat{A_{img2}}^\top},
\end{equation*}
where the adjacency matrices of two images are $\mat{A_{img1}}$ and $\mat{A_{img2}}$. The distance matrix of sublines includes the distance between descriptors from two images, which can be calculated by a matcher such as nearest neighbor.

Similar to the ensemble averaging, \eqref{eq:subline2keyline} averages the multiple sublines' distances with respect to a keyline via matrix multiplication of adjacency matrix with distance matrix. For example, the distance between keylines in \texttt{image1} and sublines in \texttt{image2} can be represented in $\mat{A_{img1}} \cdot \mat{C_{sublines}}$.
%The same principle holds on the rest equations.

%============================================================================%
\subsection{Loss Function}

We use a triplet loss function with semi-hard negative sampling strategy~\cite{Schroff2015}. The basic idea of triplet loss is to make the distance between an anchor descriptor $\bvec{a_i}$ and its matched (positive) descriptor $\bvec{p_i}$ closer and to simultaneously further the distance with a nonmatched (negative) descriptor $\bvec{n_i}$. In line segment matching, one line in one image can be matched with more than two lines in another image and this means that a single anchor line can have multiple positive lines. In our implementation, we choose the most overlapped line as the positive $\bvec{p_i}$. The overall loss function is given as:
\begin{equation}
	\label{eq:triplet_loss}
	\mathcal{L} = \frac{1}{n}\sum_{\textit{i}}^{n}\max(0, \alpha + \norm{\bvec{a_{\textit{i}}} - \bvec{p_{\textit{i}}}}^2 - \norm{\bvec{a_{\textit{i}}} - \bvec{n'_{\textit{i}}}}^2),
\end{equation}
where the semi-hard negative sample $\bvec{n'_{\textit{i}}}$ is chosen to the hard negative further away from the positive $\bvec{p_{\textit{i}}}$. We observed that semi-negative sampling helped converging loss values stably. The margin value $\alpha$ provides the capacity to increase the negative distances, and we set it to $1$.

%============================================================================%
\subsection{Implementation Details}

To detect line segments on images, we used \ac{LSD}~\cite{Gioi2010} for its high generalizability over various environments. We selected SuperPoints~\cite{DeTone2017} for our front-end descriptor map. Because its raw descriptor map is sized $H\slash8 \times W\slash8$ given an image sized $H \times W$, we set 8 to the interval of points on a line. We limited the number of point tokens greater than 2 and smaller than 21 for a subline. A line descriptor, key, query, and value in \ac{MSA} have the same dimension $\mathbf{D}=256$ as the SuperPoint. The \ac{MSA} has four heads, and the line transformers and line signature networks have $L=12$ and $M=7$ layers, respectively. Our networks contain 14M parameters, and they run at an average speed of 20 ms on an NVIDIA GTX 2080Ti GPU for 256 line descriptors in an image. It is implemented in Pytorch using the Adam optimizer with a learning rate of 0.001.

%----------------------------------------------------------------------------%
% FIGURE: flowchart
\begin{figure*}[!t]
	\centering
	\includegraphics[width=0.8\textwidth]{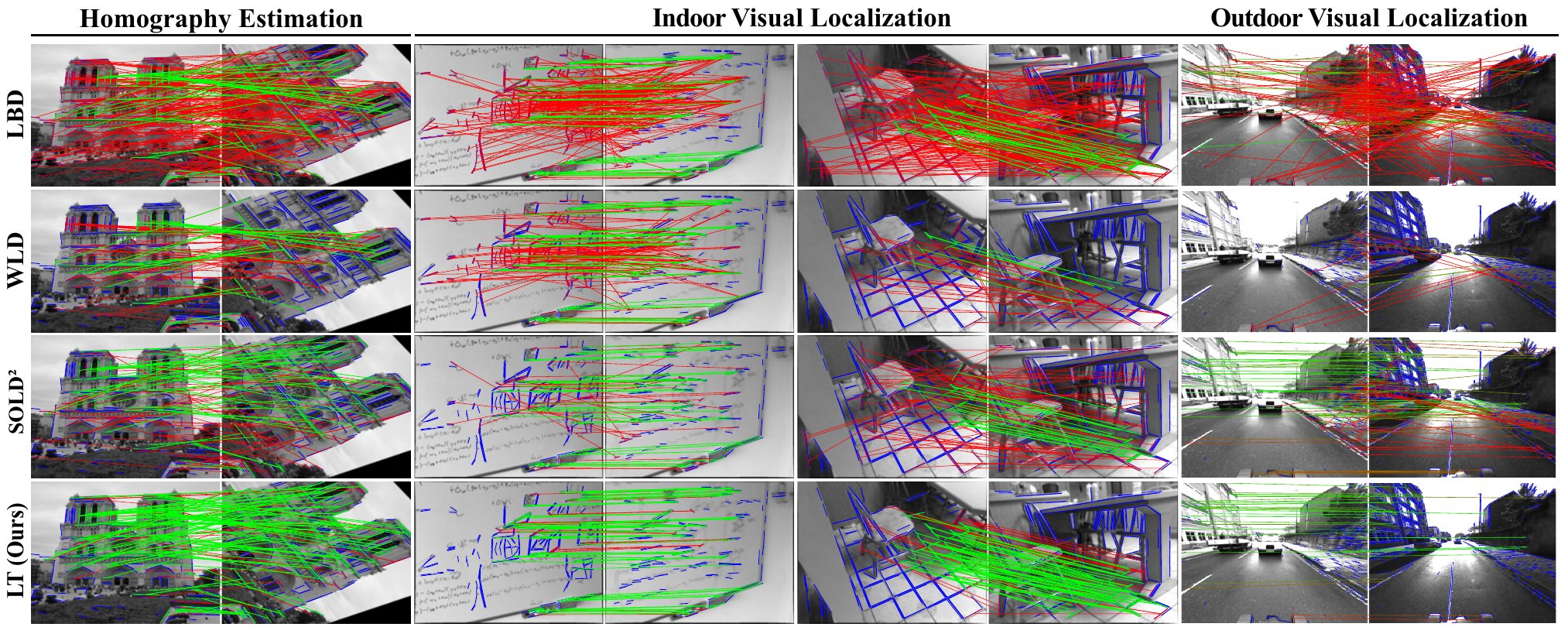}
	\caption{Qualitative line segment matches for homography estimation and visual localization. The last three columns represent blurring, viewpoint, and illumination changes. More correct matches (\textcolor{green}{green}) and fewer wrong matches (\textcolor{red}{red}) indicate better performance. Unmatched lines are colored in \textcolor{blue}{blue}. }
	\label{fig:qual}
	\vspace{-5mm}
\end{figure*}
% figure

% % FIGURE: flowchart
% \begin{figure*}[!t]
% 	\centering
% 	\includegraphics[width=\textwidth]{figures/results_homography.pdf}
% 	\caption{Qualitative line segment matches for homography estimation. Each row represents qualitative comparisons of LBD~\cite{LBD2013}, WLD~\cite{WLD2020}, SOLD$^2$~\cite{Pautrat2021}, and Line-Transformers. More correct matches (\textcolor{green}{green}) and fewer wrong matches (\textcolor{red}{red}) indicate better performance. Unmatched lines are colored in \textcolor{blue}{blue}. }
% 	\label{fig:qual_homography}
% 	\vspace{-5mm}
% \end{figure*}
% % figure

%============================================================================%
\section{Experiments}
\label{sec:exp}

We evaluate our line descriptor in terms of homography estimation and visual localization performance. For two test scenarios, we compare the proposed method against SuperPoint, handcrafted line descriptor LBD~\cite{LBD2013}, learning-based line descriptors \ac{LLD}~\cite{LLD2019}, \ac{WLD}~\cite{WLD2020}, and SOLD$^2$~\cite{Pautrat2021}. We use a \ac{NN} matcher for LBD, LLD, and WLD, and SOLD$^2$ by means of its own line matcher. We include SuperPoint as a reference for the point-based matching. We used a maximum of 512 and 1,024 points at each dataset. For lines, we extracted 256 line segments in the longer order. Further results can be seen in the video \texttt{line-as-a-visual-sentence.mp4}.

%============================================================================%
\subsection{Homography Estimation}

%----------------------------------------------------------------------------%
\subsubsection{Datasets}

For homography estimation, we use Oxford and Paris dataset~\cite{Radenovic2018}. The dataset includes 5K (Oxford) and 6K (Paris) images and we split them into the training, validatation, and test sets. For self-supervised learning, we augment images using synthetic homographies and photometric distortions~\cite{DeTone2017, Sarlin2019, Revaud2019}.

To establish ground-truth line correspondences from an image pair, we first detect line segments from both the original image and its augmented image. Then, we project two endpoints of each line onto one another using a known homography matrix. The criteria for the true correspondence are (\textit{i}) overlap existence, (\textit{ii}) reprojection error ($<$ \unit{4}{px}), and (\textit{iii}) angle difference ($<2^\circ$). The resulting true correspondences were expressed as an overlap-similarity matrix. The overlap similarity between two lines measures the ratio between overlapped line length and smaller line length as $O(L_1,L_2) = L_1 \cap L_2 / \min(L_1,L_2)$. The overlapped line length $L_1 \cap L_2$ is the distance between two middle endpoints among four endpoints of two lines. For SuperPoint, true point correspondences are defined by the point-wise reprojection errors  ($<$ \unit{4}{px}) following~\cite{Sarlin2019}.

% % FIGURE: flowchart
% \begin{figure*}[!t]
% 	\def\width{1.0\textwidth}%
% 	\centering
% 	\includegraphics[width=\width]{figures/results_indoor.pdf}
% 	\caption{Qualitative line segment matches for the visual localization dataset. Each column represents blurring, viewpoint change, and a highly non-overlapped scene. More correct matches (\textcolor{green}{green}) and fewer wrong matches (\textcolor{red}{red}) indicate better performance.} %The figure below shows that points and lines are complementary for better pose estimation, especially when keypoints are biased or just detected in small numbers.}
% 	\label{fig:qual_vl}
% 	\vspace{-5mm}
% \end{figure*}
% % figure

%----------------------------------------------------------------------------%
\subsubsection{Metrics}

We implement a line-based homography matrix estimation~\cite{Dubrofsky2008}. With the homography estimation utilizing a \ac{RANSAC} of 2,000 iterations, we compute average reprojection errors of the four image corners and report the area under the cumulative error curve (AUC) at the thresholds (5, 10, and, 20 pixels). We also compute matching precision (P) and recall (R) based on the ground-truth matches.

% TABLE
\begin{table}[!t]
\centering
\caption{Homography estimation evaluation. The best-performing values are in bold font.}
\label{tab:eval_homography}
\resizebox{\linewidth}{!}{%
\begin{tabular}{ccccccc}
\hline
\multirow{2}{*}{} & \multicolumn{3}{c}{Homography AUC}   & \multirow{2}{*}{P} & \multirow{2}{*}{R} & \multirow{2}{*}{F} \\
           & AUC@5px & AUC@10px & AUC@20px &      &      &      \\ \hline
(SuperPoint) & (38.5)    & (43.3)     & (46.3)     & (37.6) & (38.6) & (39.6) \\ \hline
LBD        & 2.2     & 7.6      & 17.5     & 20.6 & 55.2 & 30.1 \\
LLD        & 0.7     & 2.6      & 6.6      & 5.9  & 13.6 & 8.3  \\
WLD        & 16.7    & 35.2     & 54.5     & 48.0 & 51.3 & 49.6 \\
SOLD$^2$ & \textbf{31.8}    & 51.5     & 67.1     & 41.1 & 45.8 & 43.3 \\
LT (Ours)  & 29.5 & \textbf{52.1} & \textbf{69.4} & \textbf{57.7}      & \textbf{61.5}      & \textbf{59.5}      \\ \hline
\end{tabular}
}
\vspace{-5mm}
\end{table}

% TABLE

%----------------------------------------------------------------------------%
\subsubsection{Results}

\tabref{tab:eval_homography} lists the quantitative comparison. Our Line-Transformers outperform other line descriptor methods in terms of F-score by a large margin (10.1\%). Our method also outperforms every homography estimation metric except AUC under five pixels. Compared to SuperPoint, the Line-Transformers yielded more stable performance at the AUC under 10 and 20 pixels. \ac{LLD} has low performances on this dataset because it originally trained in continuous frames without large viewpoint changes.

The precision and recall are a direct and explicit measure for line-matching performance depending solely on the number of correct/wrong matches. More implicit performance could be captured from homography estimation when the performance depends on the number, distribution, and quality of the matches. In that sense, the proposed method fulfilled the quantity and quality of the reliable matches. We discuss the matching evaluation in more detail in \secref{sec:disc}.

\figref{fig:qual} shows qualitative results for the homography estimation-based line matching. Line-Transformers have better performance by producing more correct matches and fewer false matches than other line descriptors. It also shows that \ac{LBD} in particular has many incorrect matches, resulting in lower matching precision.

% FIGURE: flowchart
%\begin{figure}[!t]
%	\def\width{0.90\columnwidth}%
%	\centering
%	\includegraphics[width=\width]{figures/line_map.pdf}
%	\caption{3D line map for visual localization. Given the 3D line map, visual localization is performed by estimating the camera pose of another image with overlap.}
%	\label{fig:line_map}
%	\vspace{-5mm}
%\end{figure}
% figure

% FIGURE
\begin{figure*}[!t]
	\centering
  \begin{minipage}{0.5\textwidth}
    \subfigure[Performance difference between Line-Transformers and other descriptors.]{
	    \includegraphics[width=\textwidth]{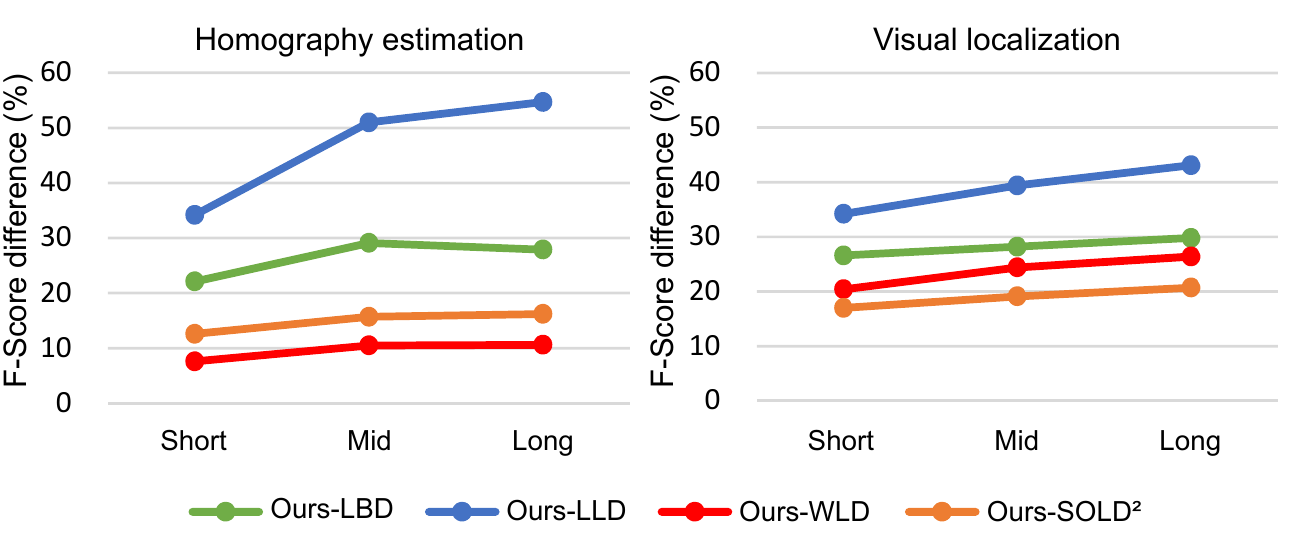} % left bottom right top
	    \label{fig:by_length}
    }
    \vspace{-2mm}
  \end{minipage}%
  \begin{minipage}{0.5\textwidth}
    \vspace{5mm}
    \subfigure[Matching performances by line length.]{
	    \label{tab:by_length}
	    \resizebox{1.0\linewidth}{1.7cm}{
			\begin{tabular}{ccccccccccc}
			\hline
			                                                                                &           & \multicolumn{3}{c}{Precision}                 & \multicolumn{3}{c}{Recall}                    & \multicolumn{3}{c}{F-score}                   \\
			                                                                                &           & Short         & Mid           & Long          & Short         & Mid           & Long          & Short         & Mid           & Long          \\ \hline
			\multirow{5}{*}{\begin{tabular}[c]{@{}c@{}}Homography\\ etimation\end{tabular}} & LBD       & 10.9          & 20.8          & 28.0          & 40.5          & 54.9          & 55.0          & 17.1          & 30.2          & 37.1          \\
			                                                                                & LLD       & 3.3           & 6.0           & 8.1           & 10.3          & 13.3          & 14.1          & 5.0           & 8.3           & 10.3          \\
			                                                                                & WLD       & 24.9          & 43.5          & 53.9          & 43.3          & 55.4          & 55.0          & 31.6          & 48.8          & 54.4          \\
			                                                                                & SOLD$^2$     & 24.3          & 39.6          & 50.0          & 29.5          & 48.5          & 47.7          & 26.6          & 43.6          & 48.8          \\
			                                                                                & LT (Ours) & \textbf{35.3} & \textbf{57.5} & \textbf{67.5} & \textbf{44.1} & \textbf{61.3} & \textbf{62.7} & \textbf{39.2} & \textbf{59.3} & \textbf{65.0} \\ \hline
			\multirow{5}{*}{\begin{tabular}[c]{@{}c@{}}Visual \\ localization\end{tabular}} & LBD       & 11.3          & 17.7          & 27.1          & 42.3          & 47.9          & 53.4          & 17.8          & 25.9          & 35.9          \\
			                                                                                & LLD       & 6.8           & 9.8           & 16.5          & 26.3          & 29.1          & 35.9          & 10.2          & 14.7          & 22.6          \\
			                                                                                & WLD       & 18.6          & 25.6          & 38.7          & 33.8          & 35.3          & 39.9          & 24.0          & 29.7          & 39.3          \\
			                                                                                & SOLD$^2$     & 24.7          & 30.8          & 45.7          & 30.8          & 40.4          & 44.4          & 27.4          & 35.0          & 45.0          \\
			                                                                                & LT (Ours) & \textbf{34.7} & \textbf{45.1} & \textbf{59.5} & \textbf{61.6} & \textbf{67.6} & \textbf{73.3} & \textbf{44.4} & \textbf{54.1} & \textbf{65.7} \\ \hline
			\end{tabular}
    }
    }
  \end{minipage}
  \caption{Performance difference by line length. The figure (a) illustrates that the overall graph has an upward trajectory, thus showing that our method performs better than other CNN-based line descriptors when line segments extend.}
  \label{fig:length_all}
	\vspace{-4mm}
\end{figure*}

% FIGURE

%============================================================================%
\subsection{Visual Localization}

Next, we evaluate line descriptors by estimating a camera pose in 3D line maps. The evaluations were performed using the ScanNet~\cite{Dai2017} and Oxford Radar RobotCar~\cite{Dan2019} datasets for indoor and outdoor experiments. For indoor environments, we trained our networks in supervised learning, and initialized them with the homography estimation's weights. Then, the weights trained using ScanNet were applied directly to outdoor localization to observe our networks' generalizability.

%----------------------------------------------------------------------------%
\subsubsection{Indoor}

The ScanNet dataset~\cite{Dai2017} includes 2.5M views of RGB-D data and 3D camera poses in 1,513 indoor scenes. We generate ground-truth line correspondences and 3D line maps. Similarly, as in the research~\cite{Dusmanu2019, Sarlin2019}, we selected sufficiently overlapping image pairs (40--80\%) by investigating the depth maps. Due to the potential inaccuracy in the depth maps, we decomposed a line into point arrays and examined the intermediate points to validate the line correspondences. Then, we randomly select 350 image pairs from each of the four overlapped scenes for the test set, a total of 1,400 image pairs. Because depth maps are not sufficiently accurate to find all ground-truth line pairs, some correct line correspondences may be counted as a false negative. Therefore, we mitigate this issue by checking scene depth quality, as well as observing the ratio between the numbers of all extracted line segments and the validated line segments during projection{\slash}unprojection procedures.

\subsubsection{Outdoor}

The Oxford Radar RobotCar dataset~\cite{Dan2019} is built upon the Oxford RobotCar dataset~\cite{Will2017}. We selected two sequences (2019-01-11-12-26-55 and 2019-01-16-13-09-37) for our reference and query sets. Then, we randomly selected 300 query images in a sequence and performed visual place recognition~\cite{Revaud2019_APGeM} to identify corresponding reference images that have a 3D line feature map. Instead of using the \ac{GPS} potential unreliable signals, we have computed the ground-truth relative pose between query and reference images using their point clouds via \ac{ICP}. In the final evaluation sets, we excluded query-reference image pairs with poor \ac{ICP} fitness. The 3D line maps were generated following procedures similar to those for our indoor evaluation. Unlike the indoor RGB-D camera, however, the projected LiDAR points are so sparse that 2D line segments can be difficult to find their corresponding depth value. To alleviate this, we utilized a depth completion~\cite{Park20}.
%The \ac{GPS} or \ac{INS} is limited to measure a precise ground-truth pose due to unstable signals. Thus, we have computed the relative pose between query-reference images using their point clouds and \ac{ICP}. In the final evaluation sets, we dropped query-reference image pairs that had poor \ac{ICP} fitness due to their inaccurate ground-truth poses. The 3D line maps were generated following procedures similar to those for our indoor evaluation. Unlike the indoor RGB-D camera, however, the projected LiDAR points are so sparse that 2D line segments can be difficult to find their corresponding depth value. To alleviate this, we utilized a depth completion~\cite{Park20} to interpolate sparse 3D points within 10 pixels.}

% FIGURE
% Please add the following required packages to your document preamble:
% \usepackage{multirow}
\begin{table}[!t]
\caption{Visual localization}
\label{tab:eval_visloc}
\centering
\resizebox{1.0\linewidth}{!}{
\begin{tabular}{cccccccccc}
\hline
  &
  \multicolumn{6}{c}{Visual Localization AUC (Indoor)} &
  \multirow{2}{*}{P} &
  \multirow{2}{*}{R} &
  \multirow{2}{*}{F} \\
            & \multicolumn{2}{c}{0.25m/10º} & \multicolumn{2}{c}{0.50m/10º} & \multicolumn{2}{c}{1.00m/10º}    &      &      &      \\ \hline
(SuperPoint)& \multicolumn{2}{c}{(83.6 PnP)} & \multicolumn{2}{c}{(86.4 PnP)} & \multicolumn{2}{c}{(86.8 PnP)} & (49.5) & (69.1) & (57.7) \\ \hline
            & PnL           & PnPL          & PnL           & PnPL          & PnL           & PnPL             &      &      &      \\
LBD         & 15.7          & 38.4          & 19.0          & 45.9          & 19.3          & 46.6             & 18.5 & 49.8 & 27.0 \\
LLD         & 12.0          & 35.8          & 16.1          & 41.6          & 16.4          & 42.5             & 10.5 & 31.8 & 15.8 \\
WLD         & 28.0          & 55.9          & 35.4          & 61.6          & 36.1          & 62.4             & 27.9 & 37.3 & 31.9 \\
SOLD$^2$    & 46.4          & 73.5          & 57.4          & 77.9          & 59.2          & 78.6             & 35.0 & 40.5 & 37.5 \\
LT (w/o LS) & 47.4          & 76.9          & 59.8          & 81.6          & 61.4          & 82.5             & 42.4 & 62.2 & 50.4 \\
LT (Ours) &
  \textbf{53.0} &
  \textbf{79.2} &
  \textbf{65.0} &
  \textbf{83.3} &
  \textbf{66.6} &
  \textbf{83.9} &
  \textbf{49.4} &
  \textbf{68.4} &
  \textbf{57.3} \\
\hline\hline
    &
    \multicolumn{6}{c}{Visual Localization AUC (Outdoor)} & \multirow{2}{*}{P} & \multirow{2}{*}{R} & \multirow{2}{*}{F} \\
             & \multicolumn{2}{c}{0.25m/2º} & \multicolumn{2}{c}{0.50m/5º} & \multicolumn{2}{c}{5.00m/10º} &      &      &      \\ \hline
(SuperPoint) & \multicolumn{2}{c}{(37.1 PnP)}   & \multicolumn{2}{c}{(63.6 PnP)} & \multicolumn{2}{c}{(90.8 PnP)}      &      &      &      \\ \hline
              & PnL           & PnPL          & PnL           & PnPL          & PnL           & PnPL          &      &      &      \\
  LBD                  & 1.9  & 4.5  & 2.6  & 18.5 & 9.1  & 59.2 & - & - & - \\
  LLD                  & 1.9  & 7.5  & 6.4  & 30.2 & 20.0 & 71.7 & - & - & - \\
  WLD                  & 9.4  & 21.5 & 29.4 & 45.7 & 54.7 & 83.4 & - & - & - \\
  SOLD$^2$             & 21.5 & 28.7 & 54.0 & 53.2 & 82.3 & 90.2 & - & - & - \\
  LT (w/o LS)          & 21.9 & 29.4 & 55.1 & 58.9 & 87.9 & 91.7 & - & - & - \\
  LT (Ours) & \textbf{26.8} & \textbf{30.9} & \textbf{57.7} & \textbf{61.1} & \textbf{90.2} & \textbf{91.3} & - & - & - \\\hline
\end{tabular}
}
\vspace{-5mm}
\end{table}

%\input{tab_visloc_results_ox}
% FIGURE

%----------------------------------------------------------------------------%
\subsubsection{Metrics}

%For validation, SuperPoint with PnP and line descriptors with \ac{PnPL} \cite{Agostinho2019} were implemented.
We report the percentage of correctly localized query images while using different thresholds (i.e., 0.25m, 10\textdegree {\slash} 0.5m, 10\textdegree {\slash} 1.0m, 10\textdegree{} for indoor and 0.25m, 2\textdegree {\slash} 0.5m, 5\textdegree {\slash} 5.0m, 10\textdegree{} for outdoor). We estimated the camera pose by \ac{PnPL}~\cite{Agostinho2019} with a \ac{RANSAC} of 20 iterations. For SuperPoint, we leveraged \ac{PnP}. We analyze the pose estimation results that use points, lines, and points-and-lines, respectively. We also report matching precision (P) and recall (R) based on the ground-truth matches at indoor.

%----------------------------------------------------------------------------%
\subsubsection{Results}

For both indoor and outdoor tests, Line-Transformers achieve the highest performance among other line descriptors in visual localization and precision-recall metrics (\tabref{tab:eval_visloc}). The qualitative results in \figref{fig:qual} illustrate that the Line-Transformers perform robustly in imaging changes such as blurring, viewpoints, and illuminations. Our method can be generalized well performing reliably using the same weights trained in ScanNet datasets.

Unlike in the homography estimation, the point-based method with \ac{PnP} outperforms all line-based methods. One of the reasons is the small number of 3D line inliers during depth validation. While the 3D feature point is directly determined by its corresponding depth pixel, some 3D line features are filtered out during depth linearity's validation in \ac{RANSAC}. Thus, within our mapping method, line-based localization is prone to performance degradation than point-based. However, as will be discussed in \textsection IV.F, the line features can complement the performance of the points, especially when point feature numbers are small or biased.

%The line-based localization outperforms SuperPoint-based localization even at the narrow threshold. Due to the low-textured indoor environments, keypoints are not detected well or are not well distributed on images. This outperformance indicates that line-based localization is advantageous in simple-structured environments where point features can be vulnerable. \figref{fig:qual} shows qualitative results for line matches in this dataset. More importantly, we validate that Line-Transformers can boost localization performance when line features are used with the point features \ac{PL-Loc}. The result shows that localization performance was enhanced with a big margin (26.2\%) under the same threshold, 0.25m, 10\textdegree.  As shown in \figref{fig:point_line}, line matches support the discovery of effective scene orientation when a small number of keypoints are detected or when the detected points are biased.

%============================================================================%
\subsection{Performance by Line Length}
\label{sec:length}

To examine the robustness of our model with respect to various line lengths, we further investigate performance comparisons against \ac{CNN}-based methods while varying line length. From each dataset, we evenly divide line segment sets into three groups (i.e., short, mid, and long) by their lengths. Each group has 33\% numbers of all line segments, and the mid-length group ranged from about 30 to 50 pixels.

As presented in \figref{tab:by_length}, the proposed Line-Transformers outperform other descriptors for all line lengths. Furthermore, as shown in \figref{fig:by_length}, the matching performance difference between Line-Transformers and other CNN-based line descriptors (\ac{LLD} and \ac{WLD}) is increased by line length. This tendency indicates that the performance of the proposed method is preserved even with longer line segments, unlike the handcrafted descriptor and CNN-based descriptor.

%As presented in \ref{tab:by_length}, every line descriptors perform better as line lengths increase although we initially expected that hand-crafted descriptor and CNN-based descriptor get worse for the longer line segments. We can also see that Line-Transformers outperform other descriptors with all line lengths, and the matching performance difference between Line-Transformers and other CNN-based line descriptors (\ac{LLD} and \ac{WLD}) gets increased by line length as shown at \ref{fig:by_length}.

\begin{figure*}[!t]
	\centering
  \begin{minipage}{0.55\textwidth}
  \subfigure[Point-and-line features]{%
    \includegraphics[width=\textwidth]{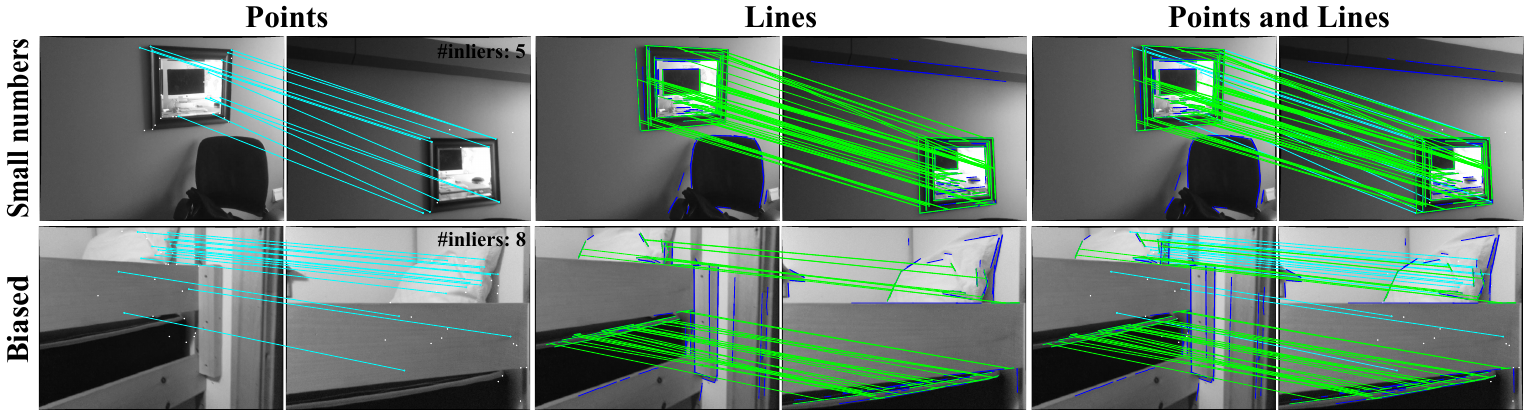}
  }
  \vspace{-2mm}
  \end{minipage}%
  \begin{minipage}{0.40\textwidth}
    %\vspace{5mm}
    \subfigure[Point-and-Line Visual localization]{
	    \label{tab:pl-loc}
	    \resizebox{1.0\linewidth}{!}{
      \begin{tabular}{ccccc}
      \hline
      \multicolumn{5}{c}{Indoor Visual Localization}                                          \\
                              & \#inliers     & 0.25m/10º     & 0.50m/10º     & 1.00m/10º     \\ \hline
      SuperPoint              & -             & 83.6          & 86.4          & 86.8          \\
      \multirow{2}{*}{PL-Loc} & \textless{}20 & 83.6          & 86.5          & 87.2          \\
                              & \textless{}5  & \textbf{84.5} & \textbf{87.4} & \textbf{88.0} \\ \hline
      Best Feature (\%)       & P/L/PL        & \multicolumn{3}{c}{61.0 / 6.3 / 32.7}         \\ \hline
      \end{tabular}
      }
      }
  \end{minipage}
  \caption{Points and lines are complementary for better localization, especially when keypoints are biased or in small numbers.}
  \label{fig:point_line}
  \vspace{-5mm}
\end{figure*}

% Please add the following required packages to your document preamble:
% \usepackage{multirow}
\begin{table}[!t]
\caption{Ablation of Line-Transformers.}
\label{tab:ablation}
\centering
\resizebox{1.0\linewidth}{!}{
\begin{tabular}{lcccccc}
\hline
\multicolumn{1}{c}{\multirow{2}{*}{Line-Transformers}} &
  \multicolumn{3}{c}{Line-based Visual   Localization} &
  \multirow{2}{*}{P} &
  \multirow{2}{*}{R} &
  \multirow{2}{*}{F} \\
\multicolumn{1}{c}{}         & 0.25m/10º & 0.50m/10º & 1.00m/10º &      &      &      \\ \hline
No Line Signature Net (LS)   & 47.4      & 59.8      & 61.4      & 42.4 & 62.2 & 50.4 \\
No Positional encoding in LS & 49.4      & 62.8      & 64.3      & 44.3 & 66.4 & 53.1 \\
No mid-point in LS           & 51.1      & 62.6      & 64.9      & 47.7 & 69.0 & 56.4 \\
No length, angle in LS       & 51.9      & 64.2      & 65.8      & 44.0 & 68.4 & 53.6 \\
\textbf{Full} &
  \textbf{53.0} &
  \textbf{65.0} &
  \textbf{66.6} &
  \textbf{49.4} &
  \textbf{68.4} &
  \textbf{57.3} \\ \hline
\end{tabular}
}
\vspace{-5mm}
\end{table}

%============================================================================%
\subsection{Discussion on Evaluation Metrics}
\label{sec:disc}

Unlike point features that assume one-to-one matching, line matching is a many-to-many problem because a line detector tends to break the same line segment into small lines differently at each image pair. For example, two non-overlapping lines may originate from a single line due to the occlusion and segmentation; thus, they should be considered the correct correspondences semantically. %Considering these properties, we discuss the advantages and limitations of evaluation metrics. In the end, we claim that line-based visual localization can be a good alternative to evaluate line descriptors.

Evaluation of line segment matches often depends on human inspections~\cite{Li2016benchmark, WLD2020, DLD2019, LBD2013}, assuming that line matches are one-to-one correspondences~\cite{LBD2013}, not many-to-many. Therefore, when a matcher finds only the closest matching pair, the precision and recall should be carefully considered cautioning many-to-many correspondences. In that sense, the precision-recall metrics may be limited because they cannot consider non-overlapping line correspondences.

Visual localization and homography estimation could be more proper metrics in this aspect. In visual localization, matching with the non-overlapped but semantically same line is also considered a good match because a \ac{PnL} algorithm does not consider the endpoint's positions to ensure the robustness for changing endpoints. Similarly, the line-based homography estimation does not consider endpoints~\cite{Dubrofsky2008} but is limited to planar scenes in the real 3D world. Hence, we found that line-based visual localization is a better alternative that can inspect both overlapped and non-overlapped line matches with large perspective changes.% in real scenes.

%============================================================================%
\subsection{Understanding Line-Transformers}
\label{sec:analysis}

\begin{figure*}[!t]
	\centering
	\includegraphics[width=0.65\textwidth]{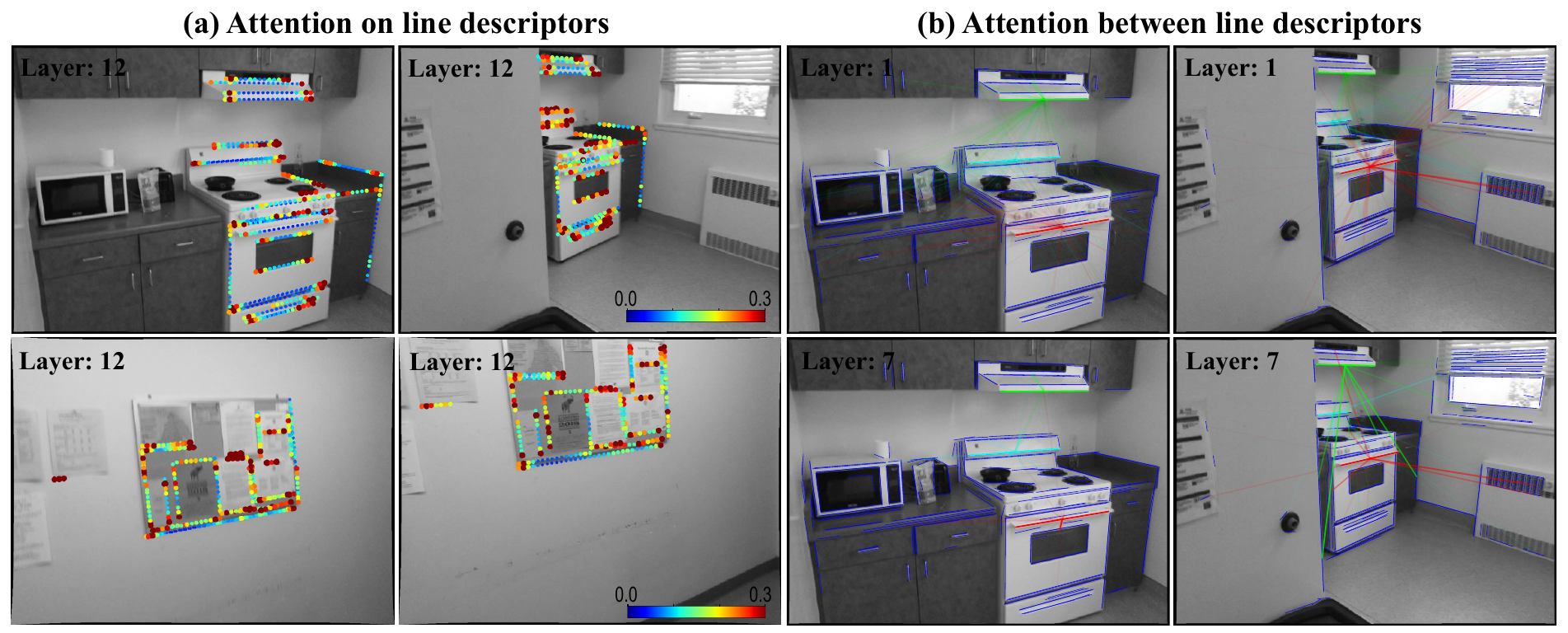}
	\caption{Visualizing attention scores. (a) Attention patterns in lines describe how much the points embeddings contribute to building line descriptors. The matched lines follow similar attention patterns. (b) Attention scores between line descriptors are initially low and widely spread, and they are gradually converged onto a small number of neighbor lines at a later layer.}
	\label{fig:vis_attn}
	\vspace{-5mm}
\end{figure*}

To examine the attention scores of Line-Transformers, we visualize the maximum attention score of each multi-heads within a line segment (\figref{fig:vis_attn}). The point embeddings on a line are not considered equally for a line descriptor, but they have their own attention patterns. We also observe that the matched lines have similar attention patterns and the line descriptors tend to refer to the endpoints of a line.

The \figref{fig:vis_attn}~(b) illustrates the attention between line descriptors in line signature networks. We observe that the attentions gradually focus on a small number of neighbor lines at a later layer. To study the quantitative effect of the line signature networks, we evaluate our models without line signature networks. As presented in \tabref{tab:eval_visloc} and \tabref{tab:ablation}, the line signature networks improve the localization results from 47.4\% to 53.0\% under the threshold of 0.25m and 10\textdegree.

% FIGURE: flowchart
% figure

% FIGURE: flowchart
%\begin{figure*}[!t]
%	\def\width{0.85\textwidth}%
%	\centering
%	\includegraphics[width=\width]{figures/point_line.pdf}
%	\caption{Point-and-Line features. The figure illustrates that points and lines are complementary for better pose estimation, especially when keypoints are biased or just detected in small numbers.}
%	\label{fig:point_line}
%	\vspace{-5mm}
%\end{figure*}
% figure

%============================================================================%
\subsection{Visual Localization Using Feature Points and Lines}
\label{sec:pl-loc}

Despite the generally better performance in \tabref{tab:eval_visloc}, the point-based localization may be deteriorated due to the small number of points (e.g., feature-less environment) or biased feature distribution as in the sample cases in \figref{fig:point_line}. This section examines how the line-based approach can enhance point-based localization in a complementary fashion.

For the analysis, we define point-based localization failure using the reprojection errors of 3D features and count the inlier when the reprojection error is less than four pixels. Then, the \ac{PL-Loc} was additionally performed for the cases when the number of inliers is less than 5 or 20 (i.e., when point-based inliers drop).

As presented in the table in \figref{tab:pl-loc}, the \ac{PL-Loc} provides additional enhancements to visual localization. More interestingly, the point outperformed over the line for 61\% of cases, which indicates the remaining 39\% of cases are with potential to be improved with lines. This also implies the proper combination of point and line would improve the overall localization performance. For example, better metrics followed by a strong model selection could be examined to complete robust PnPL in future work.

%----------------------------------------------------------------------------%
%\input{tab_vl_pl}

\section{Conclusion}
\label{sec:conclusion}

This paper presented a novel line descriptor handling variable line length effectively using attention mechanisms, inspired by \ac{NLP} tasks handling sentences and paragraphs of various lengths. We also presented a \ac{PL-Loc} pipeline leveraging keypoints and keylines simultaneously for visual localization. Our experiments demonstrated that our line descriptor achieved state-of-the-art performance in homography estimation and visual localization datasets.

%
% \input{appendix}
%\section*{ACKNOWLEDGMENT}

%\balance
\small
%\balance
\bibliographystyle{IEEEtranN}
\bibliography{string-short,references}

\end{document}